\documentclass[runningheads]{llncs}
\usepackage{graphicx}
\usepackage{subfigure}
\usepackage{comment}
\usepackage{color}
\usepackage[ruled]{algorithm2e}
\usepackage{booktabs}
\usepackage{multirow}

\usepackage{setspace}
\usepackage{amsmath,amssymb} 

\DeclareMathOperator{\lstm}{LSTM}
\begin{document}
\title{Interpretable Neural Computation for Real-World Compositional Visual Question Answering\thanks{This work was supported by NSFC (60906119) and Shanghai Pujiang Program.}}
\titlerunning{Interpretable Neural Computation for VQA}
\author{Ruixue Tang \and
Chao Ma\thanks{Corresponding author.}}

\authorrunning{R. Tang and C. Ma}
%
\institute{MoE Key Lab of Artificial Intelligence, AI Institute, Shanghai Jiao Tong University \\ \email{\{alicetang, chaoma\}@sjtu.edu.cn}}
\maketitle              
\begin{abstract}
There are two main lines of research on visual question answering (VQA): compositional model with explicit multi-hop reasoning, and monolithic network with implicit reasoning in the latent feature space. The former excels in interpretability and compositionality but fails on real-world images, while the latter usually achieves better performance due to model flexibility and parameter efficiency. We aim to combine the two to build an interpretable framework for real-world compositional VQA. In our framework, images and questions are disentangled into scene graphs and programs, and a symbolic program executor runs on them with full transparency to select the attention regions, which are then iteratively passed to a visual-linguistic pre-trained encoder to predict answers. Experiments conducted on the GQA benchmark demonstrate that our framework outperforms the compositional prior arts and achieves competitive accuracy among monolithic ones. With respect to the validity, plausibility and distribution metrics, our framework surpasses others by a considerable margin.

\end{abstract}
%
%
\section{Introduction}
The advances in deep representation learning and the development of large-scale dataset \cite{antol2015vqa} have inspired a number of pioneering approaches in visual question answering (VQA). Though neural networks are powerful, flexible and robust, recent work has repeatedly demonstrated their flaws, showing how they struggle to generalize in a systematic manner \cite{lake2017generalization}, overly adhere to superficial and potentially misleading statistical associations instead of learning true causal relations \cite{agrawal2016analyzing}. The statistical nature of these models that supports robustness and versatility is also what hinders their interpretability, modularity, and soundness.

Indeed, humans are particularly adept at making abstractions of various kinds. We instantly recognize objects and their attributes, parse complicated questions, and leverage such knowledge to reason and answer the questions. We can also clearly explain how we reason to obtain the answer. To this end, researchers consider how best to design a model that could imitate the reasoning procedure of humans while take advantages of neural networks. In particular, recent studies designed new VQA datasets, CLEVR \cite{clevr} and GQA \cite{gqa}, in which each image comes with intricate, compositional questions generated by programs, and the programs exactly represent the human reasoning procedure. They facilitate us to learn an interpretable model by the supervision of programs.

Later,  compositional models \cite{nmn,xnm,hu2017learning} show a promising direction in conferring reasoning ability for the end-to-end design by composing neural modules from a fixed predefined collection. However, the behaviors of the attention-based neural executor are still hard to explain. NS-VQA \cite{nsvqa} moves one step further proposing a neural-symbolic approach for visual question answering that fully disentangles vision and language understanding from reasoning. However, these compositional models which designed for synthetic images in CLEVR are not capable of generalizing on real-images in GQA, where real images are much more semantic and visual richness than the synthetic ones and make it challenge to inferring objects, their interactions and subtle relations. 

Drawing inspiration from NS-VQA, we further design a model for real-world visual reasoning and compositional question answering that incorporate the symbolic program execution with a vision-and-language pre-trained encoder, LXMERT \cite{lxmert}. We use neural networks as powerful tools for parsing — generating scene graph from images, and programs from questions. Next, we incorporate the symbolic program executor with the LXMERT by running the programs on the scene graphs to locate attention on object regions, which are then iteratively fed to LXMERT to finally get the answer predictions. 

The combination of symbolic program executor and the visual-linguistic encoder offers two advantages. First, the use of symbolic representation offers robustness to long, complex and compositional questions. It endows our proposed model with transparent and interpretable reasoning process. Second, the visual-linguistic encoder, LXMERT, offers great cross-modality (image and text) representations to compensate the noisy scene graph. By conducting experiments on GQA dataset, results show that our model achieves competitive accuracy among prior arts, and notably, outperforms state-of-the-art models concerning the essential qualities (i.e. validity, plausibility and distribution) which are necessarily evaluated in GQA datasets.

To summarize, our main contributions are:
\begin{itemize}
    \item We propose an interpretable framework for real-world visual reasoning and compositional question answering, where images and questions are disentangled into scene graphs and programs, respectively, and by incorporating with a vision-and-language pre-trained encoder LXMERT \cite{lxmert}, we do a soft logic reasoning over the scene graphs.
    \item We extend the seq2seq \cite{bahdanau2014neural} to predict two tokens at each step to generate the programs from questions.
    \item Our model achieves competitive accuracy among prior arts on GQA dataset, and notably, outperforms state-of-the-art models with respect to the essential qualities by a considerable margin.
\end{itemize}  

\section{Related Work}
\subsubsection{VQA}
There have been numerous prominent models that address the VQA task. By and large they can be partitioned into two groups: 1) monolithic approaches \cite{santoro2017simple,mac,lxmert}, which embed both the image and question into a feature space and infer the answer by feature fusion; 2) neural module approaches \cite{nmn,xnm,hu2017learning}, which first parse the question into a program assembly of neural modules, and then execute the modules over the image features for visual reasoning. 
\subsubsection{Scene Graph.}
This task is to produce graph representations of images in terms of objects and their relationships. Scene graphs have been shown effective in boosting several vision-language tasks \cite{johnson2015image,teney2017graph,yin2017obj2text}. However, scene graph detection is far from satisfactory compared to object detection \cite{li2018factorizable,xu2017scene,zellers2018neural}. To this end, we utilize a large-scale vision-and-language pre-trained encoder to compensate the noisy predicted scene graph.

\section{Method}
In this section, we first briefly summarize the LXMERT model (Sec. \ref{sec-lxmert}) and then describe how we incorporate symbolic program executor with it (Sec. \ref{sec-2}). 
\subsection{Preliminary: LXMERT Background}
\label{sec-lxmert}
\begin{figure}[t]
\centering
\includegraphics[width=4.8in]{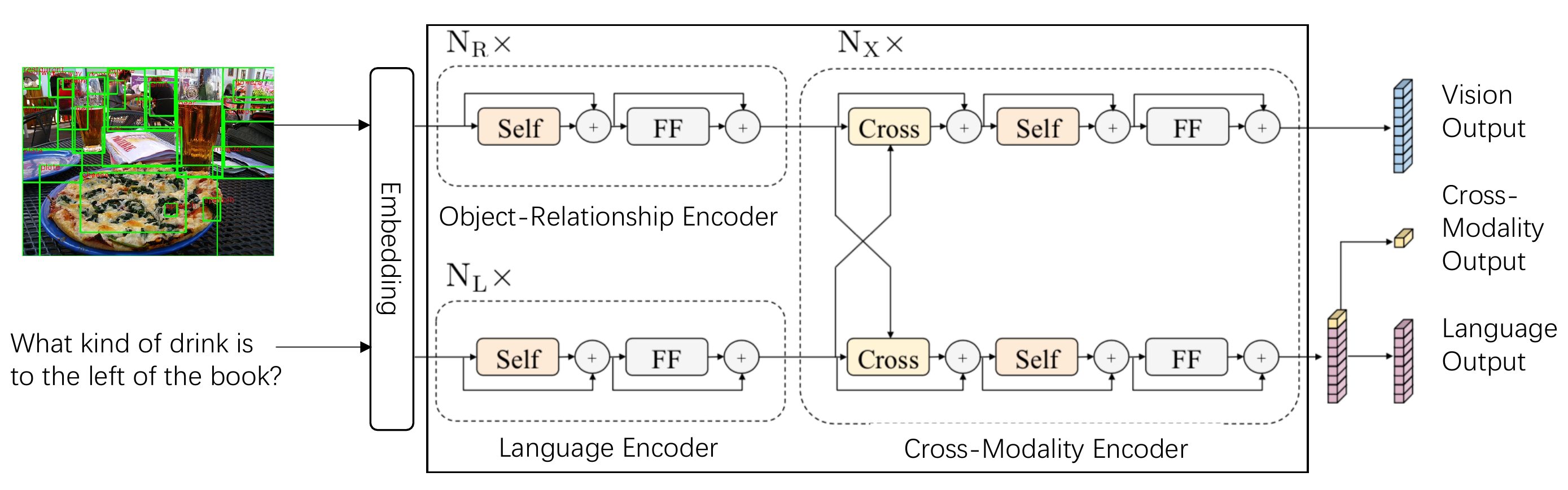}
\caption{The LXMERT model for learning vision-and-language cross-modality representations. 'Self' and 'Cross' are abbreviations for self-attention sub-layers and cross-attention sub-layers, respectively. 'FF' denotes a feed-forward sub-layer. The components in the solid block which consist of an object-relation encoder, a language encoder and a cross-modality encoder are applied in our interpretable framework introduced in the following.}
\label{lxmert}
\end{figure}

LXMERT \cite{lxmert}(Learning Cross-Modality Encoder Representations from Transformers) is a model for learning task-agnostic joint representations of image content and natural language. It is a large-scale Transformer model that consists of three encoders: an object relationship encoder, a language encoder, and a cross-modality encoder. As shown in Fig. \ref{lxmert}, LXMERT takes two inputs: an image and its related sentence (e.g., a caption or a question). Each image is represented as a sequence of objects, and each sentence is represented as a sequence of words. Via careful design and combination of these self-attention and cross-attention layers, LXMERT is able to generate language representations, image representations, and cross-modality representations from the inputs.

The \textbf{object-relationship encoder} and  \textbf{language encoder} are two transformer models and each of them only focuses on a single modality (i.e., language or vision). Each layer (left dashed blocks in Fig. \ref{lxmert}) in a single-modality encoder contains a self-attention ('Self') sub-layer and a feed-forward ('FF') sub-layer. The ('Self') sub-layer at \emph{k}-th layer could be formulated as:
\begin{align}
    \check{h}^k_i={\rm SelfAtt_{L\to L}}\Big(h_i^{k-1},\left\{h_1^{k-1}, ..., h_n^{k-1}\right\}\Big)\\
    \check{v}^k_i={\rm SelfAtt_{R\to R}}\Big(v_i^{k-1},\left\{v_1^{k-1}, ..., v_m^{k-1}\right\}\Big)
\end{align}
where $\left\{\hat{h}^k_i\right\}$ are language features and $\left\{\hat{v}^k_i\right\}$ are vision features.
We take $N_L$ and $N_R$ layers in the language encoder and the object-relationship encoder, respectively. A residual connection and layer normalization (annotated by the ‘+’ sign in Fig. \ref{lxmert}) is added after each sub-layer. Each cross-modality layer in the \textbf{cross-modality encoder} consists of one bi-directional cross-attention sub-layer, two self-attention sub-layers, and two feed-forward sub-layers. We stack $N_X$ these cross-modality layers in our encoder implementation. Inside the \emph{k}-th layer, the bi-directional cross-attention sub-layer ('Cross') is first applied, which contains two uni-directional cross-attention sub-layers: one from language to vision and one from vision to language, which are formulated as:
\begin{align}
    \hat{h}^k_i={\rm CrossAtt_{L\to R}}\Big(h_i^{k-1},\left\{v_1^{k-1}, ..., v_m^{k-1}\right\}\Big)\\
    \hat{v}^k_i={\rm CrossAtt_{R\to L}}\Big(v_i^{k-1},\left\{h_1^{k-1}, ..., h_n^{k-1}\right\}\Big)
\end{align}
A residual connection and layer normalization are added after each sub-layer as in single-modality encoders.

LXMERT is pretrained on the large scale of datasets via five diverse representative pre-training tasks: masked language modeling, masked object prediction (feature regression and label classification), cross-modality matching, and image question answering. These tasks help in learning both intra-modality and cross-modality relationships. 

\subsection{The Interpretable Neural Computation} 
\label{sec-2}
\begin{figure}[t]
\centering
\subfigure[]{
\centering
\includegraphics[width=4.8in]{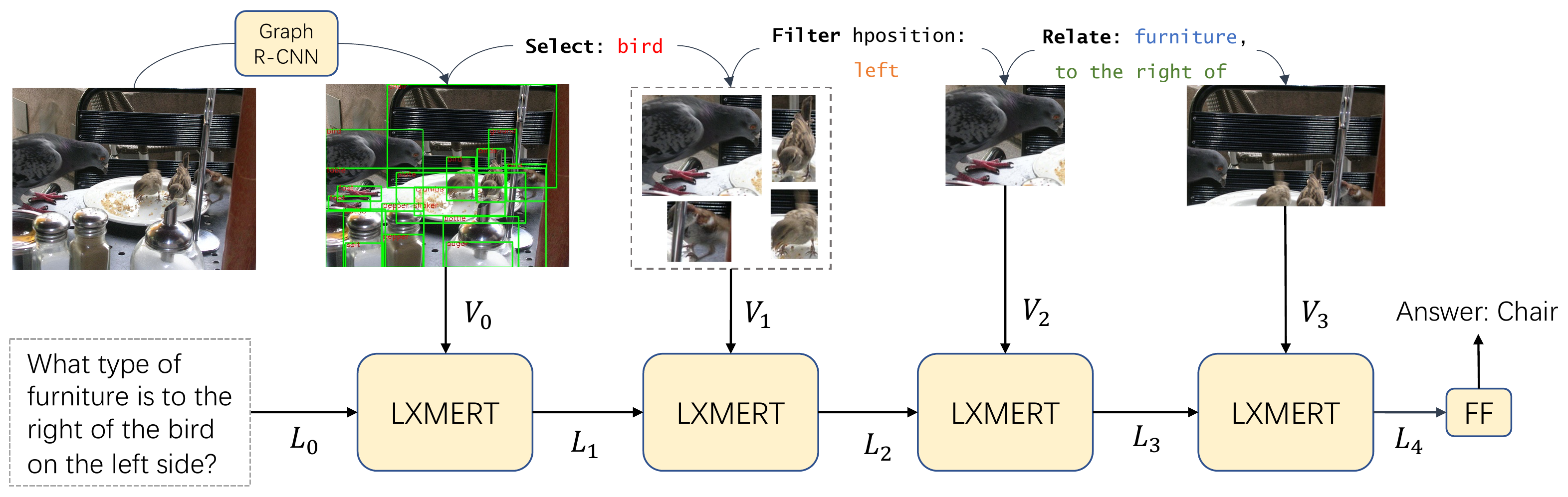}\label{gqa}
}%
\vspace{-3mm}
\quad
\subfigure[]{
\centering
\includegraphics[width=2.5in]{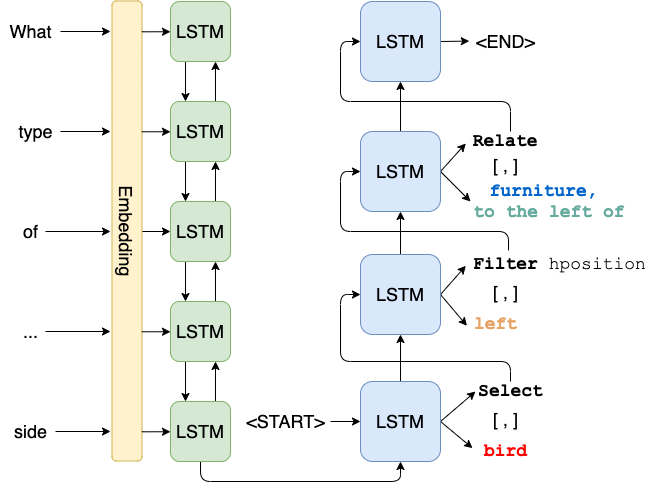}\label{lstm}
}%
\hspace{-4mm}
\subfigure[]{
\centering
\includegraphics[width=2.2in]{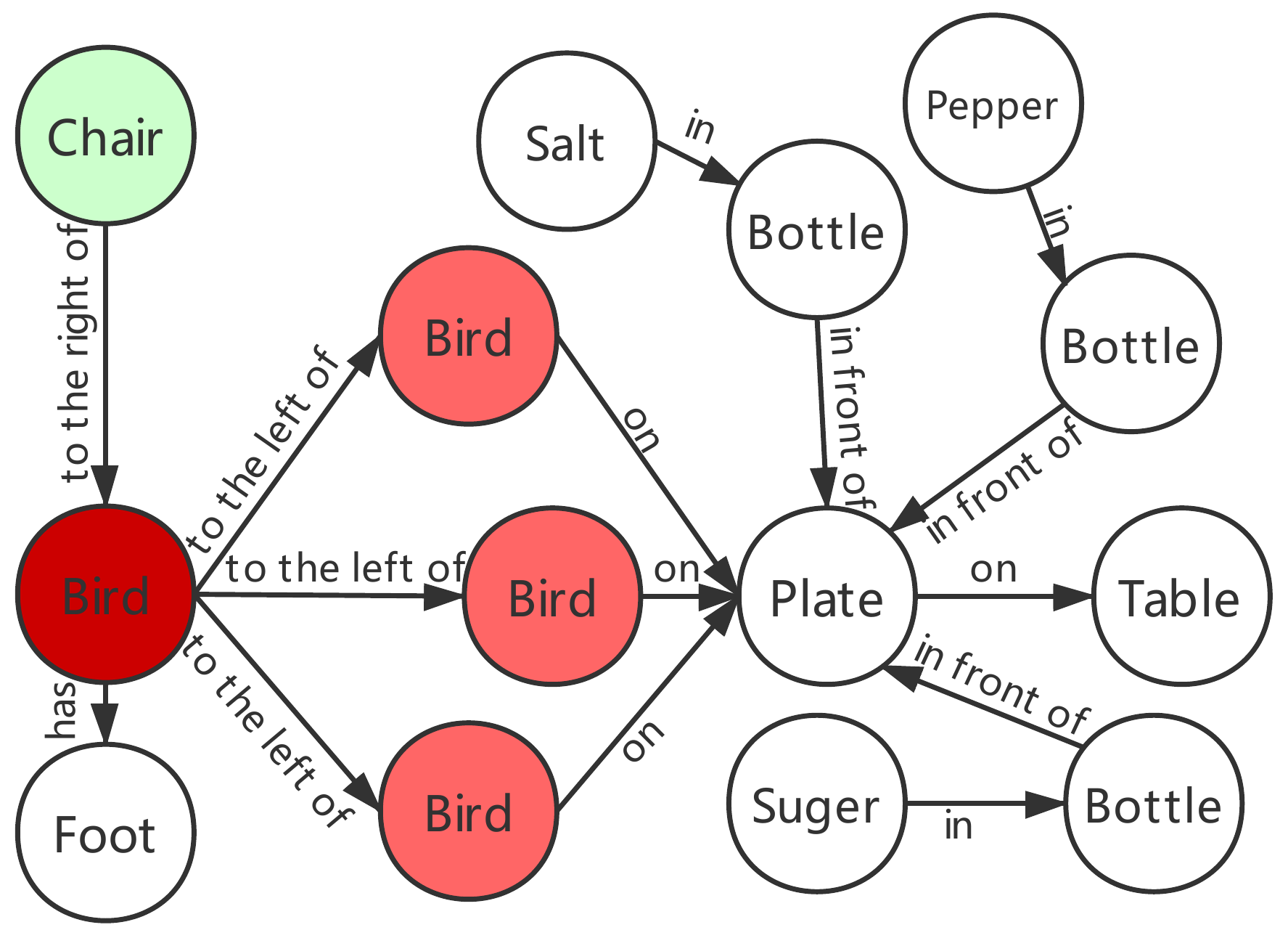}\label{scene}
}%
\vspace{-3mm}
\caption{Our interpretable framework has three components: first, a scene parser (Graph R-CNN \cite{grcnn}) that generates a scene graph (c) of each image; second, a question parser (b) that converts a question in natural language into a program; third, a neural-symbolic program executor incorporated with a neural module (LXMERT) (a), which utilizes the generated programs and scene graphs to perform soft logical reasoning.}
\label{figbig}
\vspace{-3mm}
\end{figure}

Our framework has three components: a scene parser (de-renderer), a question parser (program generator), and a symbolic program executor with LXMERT. Given an image-question pair, the scene parser de-renders the image to obtain a scene graph (Fig. \ref{scene}), the question parser generates a program from the question (Fig. \ref{lstm}), and the executor runs the program on the scene graph and passes the proposals to LXMERT to predict an answer (Fig. \ref{gqa}).

Our scene parser generates a structural and disentangled representation of the scene in the image (Fig. \ref{scene}), based on which we can perform interpretable symbolic reasoning. We use a state-of-the-art method Graph R-CNN \cite{grcnn} to generates the scene graphs, which classifies the object, its attributes and relationships.

The question parser maps an input question in natural language (Fig. \ref{lstm}) to a latent program. The program has a hierarchy of functional modules, each fulfilling an independent operation on the scene graph. Using a hierarchical program as our reasoning backbone naturally supplies compositionality and interpretability power.

The program executor takes the program output from the question parser, applies these functional modules on the scene graph of the input image, and iteratively pass the attention object regions to the LXMERT (Fig. \ref{gqa}) to predict the answers. The executable program performs purely symbolic operations on its input throughout the execution process, and by the cooperation with LXMERT, it could not only compensate the noisy scene graph but also endow the reasoning process with interpretability. In the following section, we introduce the three components in detail.

\subsubsection{Scene Parser}
The scene parser de-renders the image to obtain a structural scene representation, named scene graph (see Fig. \ref{scene}). It generates a number of region proposals, and for each region, classifies the objects, attributes and their relationships. 
We use Graph R-CNN \cite{grcnn} to generate scene graphs for each image. Graph R-CNN is the prior art on scene graph generation, which is both effective and efficient at detecting objects and their relations in images. It utilizes Faster R-CNN \cite{frcnn} to localize a set of object regions, and a relation proposal network (RePN) to learn to efficiently compute relatedness scores between object pairs which are used to intelligently prune unlikely scene graph connections (as opposed to random pruning in prior work). Then, given the resulting sparsely connected scene graph candidate, it applies an attentional graph convolution network (aGCN) to propagate higher-order context throughout the graph — updating each object and relationship representation based on its neighbors. As for attributes of each object, we additionally train a simple CNN to classify them.
\subsubsection{Question Parser}
In GQA, each question could be represented as a program which is viewed as a sequence of functions $P=f_0, f_1,...f_T$. Each function then sequentially executes on the output of the previous one to finally obtain the answer. There are 12 different categories of functions based on their coarse-grained functionality (e.g., “relate, verify, filter, choose, or”). In total, there are 139 fine-grained functions (e.g., “verify material, filter color, choose left$|$right”). For example, in Fig. \ref{figbig}, $f_2$ is {\tt Relate: furniture, to the right of, [1]}, the functionality of which is to find a furniture which is to the right if the objects returned by $f_1$: {\tt Filter hposition: left [0]}. {\tt Relate} and {\tt Filter} are called ``operation''. {\tt [1]} is the ``dependency'' which implies that function $f_2$ depends on the second execution results and {\tt furniture, to the right of} is the ``arguments''.

Our question parser is to generate operation and arguments of each function from question sentence (dependency could be inferred from the generated function sequence, so we need not predict it). We apply an attention-based sequence to sequence (seq2seq) model with an encoder-decoder structure similar to that in \cite{bahdanau2014neural}. The encoder is a bidirectional LSTM \cite{lstm} that takes as input a question of variable lengths and outputs an encoded vector $e_i$ at time step $i$ as
\begin{align}
e_i =[e_i^F, e_i^B]
\end{align}
\vspace{-20pt}
\begin{align}
e_i^F, h_i^F =\lstm(\Phi_E(x_i),h_{i-1}^F)\quad e_i^B, h_i^B=\lstm(\Phi_E(x_i),h_{i+1}^B)
\end{align}
Here $\Phi_E$ is the jointly trained encoder word embedding. $(e_i^F, h_i^F)$, $(e_i^B, h_i^B)$ are the outputs and hidden vectors of the forward and backward networks at time step $i$. The decoder is a similar LSTM that generates a vector $q_t$ from the previous token of the output sequence. In this paper, we extend the decoder LSTM to have two heads where output operation tokens and arguments tokens, respectively (see Fig. \ref{lstm}). We concatenate the embedding of operation $o_{t-1}$ and arguments $g_{t-1}$ as the input of LSTM at step $t$.  $q_t$ is then fed to an attention layer to obtain a context vector $c_t$ as a weighted sum of the encoded states via
\begin{align}
q_t =\lstm([\Phi_D(o_{t-1}),\Phi_D(g_{t-1})]), \quad \alpha_{t_i}\propto\exp(q_t^\top W_Ae_i), \quad c_t=\sum_i\alpha_{t_i}e_i
\end{align}
where $\Phi_D$ is the decoder word embedding. For simplicity we set the dimensions of vectors $q_t$, $e_i$ to be the same and let the attention weight matrix $W_A$ to be an identity matrix. Finally, the context vector, together with the decoder output, is passed to a fully connected layers with softmax activation to obtain the distribution for the predicted token $o_t\sim{\rm softmax}(W_O[q_t,c_t])$ and $g_t\sim{\rm softmax}(W_G[q_t,c_t])$. Both the encoder and decoder have two hidden layers with a 256-dim hidden vector. We set the dimensions of both the encoder and decoder word vectors to be 300.
\subsubsection{The Soft Logic Program Execution}
Previous neural-symbolic based work \cite{nsvqa} implements the program executor as a collection of deterministic, generic functional modules in Python, designed to host all logic operations on the scene graphs of synthetic images. However, the state-of-the-art scene graph generation of real images is far from practical to support all logical operations (the actual R@20 results of scene graph generation methods are around 25\% while the desired are near 90\%). Therefore, we apply a soft logic reasoning process that incorporates the symbolic program executor into a neural module (LXMERT \cite{lxmert}) which learns a strong vision-and-language representation to compensate the noisy scene graphs and on the other side, the program executor endows the LXMERT with the interpretability and sound attention regions.

\paragraph{The Selection of Functions.}
The symbolic program executor performs purely symbolic operations on its input.
For example, given a scene graph (see Fig. \ref{scene}), {\tt Select: bird} is to find the bounding boxes that contain bird – the four red circles. {\tt Filter hposition: left [0]} is to pick the one that locates on the left side (the darker red circle). {\tt Relate: furniture, to the right of, [1]} is to find the furniture that is to the right of the picked bird – the green circle.
The functions output values could be \emph{List of Objects}, \emph{Boolean} or \emph{String} (we call these function sets $\mathcal{F}_O$, $\mathcal{F}_B$ and $\mathcal{F}_S$, respectively, for simplicity), where \emph{Object} specifically refers to the detected bounding box and \emph{String} could refer to object name, attributes, relations, etc. Generally, 
$\mathcal{F}_O$ would re-locate the attention of the object regions while $\mathcal{F}_B$ and $\mathcal{F}_S$ would not, and $\mathcal{F}_O$ is always at the front of the function sequence and followed $\mathcal{F}_B$ and $\mathcal{F}_S$. To help LXMERT locate attention on proper regions, $\mathcal{F}_O$ is chose to incorporate with LXMERT and others are ignored. For example, in Fig. \ref{figbig}, the functions that with operation {\tt Select}, {\tt Filter} and {\tt Relate} are selected to perform re-locating regions in our framework. 

\paragraph{Input Embeddings.}
We need to convert the inputs of LXMERT (i.e., an image and a question) into two sequences of features: word-level question embeddings and object-level image embeddings. The question embeddings are $\left\{h_i\right\}$: 
\begin{align}
\hat{w}_i = {\rm WordEmbed}(w_i), \quad \hat{u}_i = {\rm IdxEmbed}(i), \quad h_i = {\rm LayerNorm}(\hat{w}_i +\hat{u}_i)
\end{align}
where $w_i$ is a word in question sentence $\left\{w_1, ..., w_n \right\}$, $i$ is $w_i$'s absolute position in the sentence. The object-level image embeddings are $\left\{v_j\right\}$: 
\begin{align}
\hat{r}_j = {\rm LayerNorm}(W_Fr_j+b_F), ~ \hat{p}_j= {\rm LayerNorm}(W_Pp_j+b_P), ~ v_j = (\hat{r}_j +\hat{p}_j)/2
\label{ob}
\end{align}
where each object $o_j$ is represented by its position feature (i.e., bounding box coordinates) $p_j$ and its 2048-dimensional region-of-interest (RoI) feature $r_i$. We learn a position-aware embedding $v_j$ by adding outputs of 2 fully-connected layers. 
\paragraph{The Layout}
As depicted in Fig. \ref{gqa}, firstly, object-level image embeddings of all the objects in each image($V_0=\left\{v_j\right\}$) and question embeddings ($L_0=\left\{h_i \right\}$) are fed to LXMERT. 
Then, the function $f_t\in \mathcal{P}_O$ selects a list of objects at every step $t\in \left\{0,...,T\right\}$ via our program executor, and the embeddings of selected object $V_{t+1}$, together with previous language representation output of LXMERT $L_{t+1}$, are passed to LXMERT to output the next language representation $L_{t+2}$. These could be formulated as:
\begin{align}
L_0 = \left\{h_i \right\}, ~V_0=\left\{v_j\right\}, ~V_{t+1}={\rm ImEmb}({\rm Exe}(f_{t}, V_{t})),~ L_{t+1} = {\rm LXMERT}(V_{t}, L_{t})
\end{align}
where ${\rm ImEmb}()$ is the object-level image embeddings formulated in Eq. \ref{ob}, ${\rm Exe}()$ is the symbolic executor. We only pass the hidden representation via the language output of LXMERT. 
Note that the LXMERT modules share the parameters. Finally, the language output at the last step is passed to a feed-forward network with softmax activation to obtain the distribution for the predicted answers.

\section{Experiment}
In this section, we conduct the following experiments: we evaluate our proposed model and its components on GQA v1.1 dataset \cite{gqa}.
\subsection{Dataset}
We demonstrate the value and performance of our model on the “balanced-split” of GQA v1.1, which contains 1M questions over 140K images with a more balanced answer distribution.
Compared with the VQA v2.0 dataset \cite{antol2015vqa}, the questions in GQA are designed to require multi-hop reasoning to test the reasoning skills of developed models. Compared with the CLEVR dataset \cite{clevr}, GQA greatly increases the complexity of the semantic structure of questions, leading to a more diverse function set. The real-world images in GQA also bring in a bigger challenge in visual understanding. Following \cite{gqa}, the main evaluation metrics used in our experiments are accuracy, plausibility, validity and distribution.
\subsection{Implementation Details}
\subsubsection{Parser.}
There are 1702 object classes and 310 relation classes in GQA dataset, and it has a long tail. Therefore, to train the Graph R-CNN, we only use the most frequent 500 object categories which account for 93.1\% of the total instances and 50 relation categories which account for 95.9\%. We use Faster R-CNN \cite{frcnn} associated with ResNet101 \cite{resnet} as the backbone based on the PyTorch re-implementation. During training, the number of proposals from RPN is 256. For each proposal, we perform ROI align pooling, to get a 7$\times$7 response map, which is then fed to a two-layer MLP to obtain each proposal’s representation. we perform stage-wise training — we first pre-trained Faster R-CNN for object detection, and then fix the parameters in the backbone to train the scene graph generation model. SGD is used as the optimizer, with initial learning rate 1e-2 for both training stages. For question parser, we train with learning rate $7\times10^{-4}$ for 20,000 iterations. The batch size is fixed to be 64. 
\subsubsection{Program Executor with LXMERT.}
For the LXMERT module, we set the number of layers $N_L$, $N_X$, and $N_R$ to 9, 5, and 5 respectively.
We initialize the weights of LXMERT using the model pre-trained on the large aggregated dataset\footnote{https://github.com/airsplay/lxmert}. To train the executor, we use a learning rate of $1e-5$, a batch size of 32, and fine-tune the model from pre-trained parameters for 4 epochs.
\subsection{Results}
\subsubsection{VQA}
We compare our performance both with baselines, as appear in \cite{gqa}, as well as with other prior arts of VQA model. Apart from the standard accuracy metric and the more detailed type-based diagnosis (i.e. Binary, Open), we get further insight into reasoning capabilities by reporting three more metrics \cite{gqa}: \textbf{Validity}, \textbf{Plausibility} and \textbf{Distribution}. 
The validity metric checks whether a given answer is in the question scope, e.g. responding some color to a color question. The plausibility score goes a step further, measuring whether the answer is reasonable, or makes sense, given the question (e.g. elephant usually do not eat, say, pizza). The distribution score measures the overall match between the true answer distribution and the model predicted distribution (for this metric, lower is better). 
As Table \ref{tab1} shows, our model achieves competitive accuracy among published approaches. Notably, our model outperforms state-of-the-art models (especially, LXMERT) with respect to the three metrics, which indicates that our model has more a comprehensive understanding of questions and does not learn from the data bias. NMN \cite{nmn} is also a compositional method that uses the program supervision, but lack of cross-modality learning. 
Its poor performance implies the complex content in real images and question induces challenge to multi-hop reasoning. The strong image and language representations play a crucial role in our framework.
More visualizations of the reasoning process in our method are provided in the supplementary material. 
\setlength{\tabcolsep}{4.5pt}
\begin{table}[t]
\small
\begin{center}
\caption{VQA results for single-model settings. The models with * are submissions to the GQA server but without published paper. For distribution score, lower is better.}
\label{tab1}
\begin{tabular}{lcccccc}
\toprule
Model   &  Binary  & Open & Validity &Plausibility & Distribution$\downarrow$& Accuracy \\
\midrule
Local Prior \cite{gqa} & 47.90 &16.66& 84.33 &84.31 &13.98 &31.24\\
Language \cite{gqa}& 61.90 &22.69 &96.39& 87.30 &17.93&41.07\\
Vision \cite{gqa}& 36.05& 1.74 & 35.78 &34.84 &19.99&17.82\\
Lang+Vis \cite{gqa}& 63.26 & 31.80 & 96.02&84.25 &7.46 &46.55\\
BottomUp \cite{butd} & 66.64 & 34.83 &96.18 &84.57 &5.98&49.74\\
MAC \cite{mac} & 71.23 & 38.91 &96.16&84.48 &5.34 &54.06\\
NMN \cite{nmn} & 72.88& 40.53&-&-&-&55.70 \\
LCGN \cite{lcgn} & 73.77 & 42.33 & 96.48&84.81&4.70 &57.07\\
BAN \cite{ban} & 76.00 & 40.41 & 96.16&85.58 &10.52 &57.10\\
SK T-Brain* & 77.42 & 43.10 & 96.26&85.27 &7.54 &59.19\\
PVR* & 77.69 & 43.01 & 96.45&84.53 &5.80 &59.27\\
LXMERT \cite{lxmert} & 77.16 & 45.47 & 96.35 &84.53 &5.69&60.33\\
\hline
Ours  & 75.62 & 41.07 & \textbf{96.87} & \textbf{87.94} & \textbf{2.72} &58.50\\
\bottomrule
\end{tabular}
\vspace{-5mm}
\end{center}
\end{table}

\setlength{\tabcolsep}{10pt}
\begin{table}[t]
\small
\begin{center}
\caption{The accuracy (\%) of our question parser and symbolic executor. Program Acc. represents the accuracy of generated program, which is evaluated by the accuracy of operation token, arguments token and the function (It is positive when both operation and arguments in a function are correct). Executor Acc. represents the accuracy of the answers obtained by our deterministic part of program executor executed on the ground-truth scene graph, by using ground-truth (G.T.) and generated (Gen.) program.}
\label{tab2}
\begin{tabular}{l ccc | cc}
\toprule
\multirow{2}{*}{Data Split} & \multicolumn{3}{c|}{Program Acc.} &\multicolumn{2}{c}{Executor Acc.} \\
\cmidrule(r){2-6} 
& Operation   &  Arguments  & Function & G.T.&Gen. \\
\midrule
Testdev & 97.31  & 82.52  & 81.76&- &-\\
Val & 99.01 & 83.20 & 83.06&97.71 &91.46\\
\bottomrule
\end{tabular}
\vspace{-3mm}
\end{center}
\end{table}

\subsubsection{The Performance of Each Components.}
Since each component in our framework has essential impacts on final prediction, we also report the individual performance of them.
Our question parser could reach the accuracy of 83.06\% with respect to the function instances, specifically, 99.01\% for ``operation'', 83.2\% for ``arguments'' on val set (see Table \ref{tab2}). It implies our generated programs could recover the semantics of questions and provide sound interpretable reasoning skill.

We evaluate the scene graph generation via widely adopted metrics {\tt SGDet}@K – the Recall@K of predicting objects and their relationships from scratch. We obtain 17.43, 22, 24.61 on {\tt SGDet}@20, @50, @100, respectively.

We evaluate our deterministic symbolic program executor by applying it on ground-truth scene graph. Due to the incomplete and ambiguous annotations of scene graphs in GQA, the ground truth programs execute on the ground-truth scene graph results in a VQA accuracy of 97.71\%. It revealing the performance upper-bound of models for GQA dataset. The accuracy of answers obtained by generated programs is 91.46\%. These results suggest that the noisy visual recognition hinders the ``high-level'' logical reasoning in our model leading to the marginal decrease on the accuracy, but our interpretability enhanced design indeed helps our model to predict more valid and plausible answers.

\section{Conclusion}
In this paper, we have proposed an interpretable model for real-world compositional visual question answering, which combines the compositional and monolithic model to leverage the merits of both. In our method, images and questions are disentangled into scene graphs and programs, and a symbolic program executor runs on them with full transparency to locate the attention regions, which are then iteratively passed to a visual-linguistic pre-trained encoder to predict answers. The proposed model can explain its reasoning steps with a sequence of image attentions because of the symbolic nature of the execution. Experimental results demonstrate that our method not only achieves competitive accuracy among previous works, but also could predict the answers with more validity and plausibility.

%

%
%
%
 \bibliographystyle{splncs04}
 \bibliography{48}

\end{document}